\def\year{2020}\relax
\documentclass[letterpaper]{article} 
\usepackage{aaai20}  
\usepackage{times}  
\usepackage{helvet} 
\usepackage{courier}  
\usepackage[hyphens]{url}  
\usepackage{graphicx} 
\usepackage{graphicx}  
\usepackage{researchpack}

\newcommand{\prob}{\ensuremath{\mathbb{P}}}

\usepackage{researchpack}
\usepackage{xcolor}

\title{Does Symbolic Knowledge Prevent Adversarial Fooling?}
\author{
    Stefano Teso\\
    University of Trento\\
    stefano.teso@unitn.it
}

\begin{document}
\maketitle

\begin{abstract}

    Arguments in favor of injecting symbolic knowledge into neural architectures abound.  When done right, constraining a sub-symbolic model can substantially improve its performance, sample complexity, interpretability, and can prevent it from predicting invalid configurations.  Focusing on deep probabilistic (logical) graphical models -- i.e., constrained joint distributions whose parameters are determined (in part) by neural nets based on low-level inputs -- we draw attention to an elementary but unintended consequence of symbolic knowledge:  that the resulting constraints can propagate the negative effects of adversarial examples.

\end{abstract}

\section{Introduction}

Deep probabilistic (logical) graphical models (dPGMs) tie together a sub-symbolic level that processes low-level inputs with a symbolic level that handles logical and probabilistic inference, see for instance~\cite{de2019neuro}.  The two levels are often implemented with $k$ neural networks and one probabilistic (logical) graphical models, respectively. Prominent examples of dPGMs include DeepProbLog~\cite{manhaeve2018deepproblog} and ``neural'' extensions of Markov Logic~\cite{lippi2009prediction,marra2019neural}.
In this preliminary investigation, we show with a concrete toy example that fooling a single neural network with an adversarial example~\cite{szegedy2013intriguing,biggio2018wild} can corrupt the state of multiple output variables.  We develop an intuition of this phenomenon and show that it occurs despite the model being probabilistic and regardless of whether the symbolic knowledge is factually correct.

\section{Deep Probabilistic-Logical Models}

We restrict ourselves to deep Bayesian networks (dBNs), i.e., directed dPGMs stripped of their logical component.  (Our arguments do transfer to other dPGMs and deep statistical-relational models too.)  These models are Bayesian networks where some conditional distributions are implemented as neural networks feeding on low-level inputs, and (roughly speaking) correspond to ground DeepProbLog models.

Let us illustrate them with a restricted version of the addition example from~\cite{manhaeve2018deepproblog}: the goal is to recognize the digits $x_1$, $x_2 \in \{1, \ldots, 4\}$ appearing in two MNIST images $z_1$ and $z_2$, knowing that the digits satisfy the constraint $\varphi = (x_1 + x_2 = 5)$.  Notice that the only valid predictions are $(1, 4)$, $(2, 3)$, $(3, 2)$, $(4, 1)$.

Let $x = (x_1, x_2)$ and $z = (z_1, z_2)$.  Our dBN for this problem defines a joint distribution $\prob(x \,|\, z; \varphi)$ built on the conditionals $\prob(x_1 \,|\, z_1)$, $\prob(x_2 \,|\, z_2)$ and on $\varphi$.  In particular, the probability of the event $X_i = x_i$ is implemented as a ConvNet with a softmax output layer applied to $z_i$.  The dBN is consistent with the symbolic knowledge $\varphi$ in that it ensures that the joint distribution satisfies $\prob(x \,|\, z; \varphi) = 0$ for all $x \not\models \varphi$.  This is achieved by taking an unconstrained joint distribution $\prob(x \,|\, z) = \prod_i \prob(x_i \,|\, z_i)$ and constraining it:
\[
    \prob(x \,|\, z; \varphi) = \prob(x \,|\, z) \Ind{x \models \varphi} / Z
\]
Here $Z = \sum_{x \models \varphi} \prob(x \,|\, z)$ is a normalization constant and the sum runs over all $x$'s consistent with $\varphi$.  A joint prediction is obtained via maximum a-posteriori (MAP) inference~\cite{koller2009probabilistic}:
\[
    \textstyle
    F(z) = \argmax_x \prob(x \,|\, z; \varphi)
    \label{eq:F}
\]
If no symbolic knowledge $\varphi$ was given, the most likely outputs would simply be $(f_1(z_1), f_2(z_2))$, where:
\[
    \textstyle
    f_i(z_i) = \argmax_{x_i} \prob(x_i \,|\, z_i)
    \label{eq:f}
\]
Finally, we use the same ConvNet for both images, and let $f\!=\!f_1\!=\!f_2$.

\section{Adversarial Examples and Constraints}

Consider a pair of images $z$ representing a $1$ and a $4$, respectively, and let the ConvNet output the following conditional probabilities:
\begin{align}
    \prob(X_1 \,|\, z_1)
        & \textstyle = (0.9, 0.1, 0, 0)
    \\
    \prob(X_2 \,|\, z_2)
        & \textstyle = \left(\frac{1}{4} - \frac{\epsilon}{3}, \frac{1}{4} - \frac{\epsilon}{3}, \frac{1}{4} - \frac{\epsilon}{3}, \frac{1}{4} + \epsilon\right)
\end{align}
for some small $\epsilon$, e.g., $0.001$.  Although the second image is rather uninformative, the unconstrained dPGM gets both digits right, with joint probability $\approx 0.226$ (by Eq.~\ref{eq:f}) and so does the constrained classifier, with probability $\approx 0.9$ (Eq.~\ref{eq:F}).  In this case, the symbolic knowledge boosts the confidence of the model, a desirable and expected result.

Now, perturbing $z_i$ by $\delta_i$ shifts the conditional distribution output by the ConvNet from $\prob(x_i \,|\, z_i)$ to $\prob(x_i \,|\, z_i + \delta_i)$ and hence changes the probabilities assigned to the possible outcomes $X_i$.  Intuitively, a perturbation is adversarial if it is at the same time imperceptible and it forces MAP inference to output a wrong configuration.  In other words, assuming that $z_i$ is classified correctly, $z_i + \delta_i$ is adversarial if $f(z_i + \delta_i) \ne f(z_i)$ and $\|\delta_i\|$ is ``small'' for some norm $\|\cdot\|$.

It is well known that neural networks are often susceptible to rather eye-catching adversarial perturbations that can alter their output by arbitrary amounts~\cite{szegedy2013intriguing,biggio2018wild}.  Thus it is not too far fetched to imagine a perturbation $\delta_1$ that induces the following conditional distribution on the first digit:
\[
    \prob(X_1 \,|\, z_1 + \delta_1) = (0.1, 0.9, 0, 0)
\]
Now, it can be readily verified that this perturbation forces the unconstrained dBN to predict $(2, 4)$ with joint probability $\approx 0.226$ (which is symmetrical to the above case).  Clearly this model is fooled by the adversarial image into making a mistake on $x_1$, but the damage is limited to the first digit: \emph{$x_2$ is still predicted correctly}.

However $(2, 4)$ does violate the symbolic knowledge $\varphi$, while the constrained dBN is forced to output a valid prediction, namely the most likely configuration out of $\{(1, 4), (2, 3), (3, 2), (4, 1)\}$.  Given the above conditional distributions and $\epsilon$, the constrained dBN outputs $(2, 3)$ with probability $\approx 0.9$.  This prediction is definitely consistent with $\varphi$, but now \emph{both digits are classified wrongly}.

\section{Discussion}

The toy example above illustrates the perhaps elementary but seemingly neglected fact that symbolic knowledge can propagate the negative effects of adversarial examples.  This occurs because the model trades off predictive loss in exchange for satisfying a hard constraint.

While our example is decidedly toy, it is easy to see that the same phenomenon could occur in relevant sensitive applications.  The phenomenon is also likely to transfer to undirected dPGMs like deep extensions of  Markov Logic Networks~\cite{lippi2009prediction,marra2019neural}.

We make a couple of important remarks.  First, depending on the structure of the symbolic knowledge, fooling a single neural networks in the dPGM may perturb any subset of output variables.  Thus, seeking robustness of a single network is not enough and all $k$ networks must be robustified.  Second, this may not be enough either: if an adversary manages to fool a robustified neural network -- even by random luck -- the effects of fooling will still cascade across the model.  Thus the dPGM \emph{as a whole} must be made robust, in the sense that all CPTs appearing in it -- not only the ConvNets -- must be made robust.  Finally, it may be the case that access to the symbolic knowledge might help attackers in designing minimal targeted attacks that induce any target variable.

Adversarial examples in dPGMs can be understood through the lens of sensitivity analysis for directed~\cite{chan2002numbers} and undirected probabilistic graphical models~\cite{chan2005sensitivity}; see especially~\cite{chan2006robustness}.  These works show how to constrain a probabilistic graphical model to ensure that the probabilities of different queries are sufficiently far apart.  These constraints could be injected into standard adversarial training routines for neural networks to encourage global robustness of the dPGM.
Of course, robust training of complex dPGMs is likely to be computationally challenging.  Algebraic model counting in the sensitivity semiring might prove useful in tackling this computational challenge~\cite{kimmig2017algebraic}.

\bibliographystyle{aaai}
\bibliography{paper}
\end{document}